\documentclass[runningheads]{llncs}

\usepackage{eccv}

\usepackage{eccvabbrv}

\usepackage{graphicx}
\usepackage{booktabs}

\usepackage[accsupp]{axessibility}  

\usepackage{hyperref}

\usepackage{orcidlink}

\makeatletter  
\def\@fnsymbol#1{\ensuremath{\ifcase#1\or  
  \textrm{*}\or              
  \ddagger\or       
  \dagger\or       
  \mathparagraph\or 
  \|\or  
  **\or  
  \dagger\dagger\or  
  \ddagger\ddagger\else\@ctrerr\fi}}  
\makeatother 
\usepackage{graphicx}
\usepackage{booktabs}
\usepackage{makecell}
\usepackage{multirow}
\usepackage{bbding}
\usepackage{xcolor} 
\usepackage{pifont}
\usepackage{colortbl}
\usepackage{hyperref}
\usepackage{tcolorbox}
\tcbuselibrary{breakable}
\usepackage{pgffor}
\usepackage{booktabs}  
\newcommand{\imgToken}[1]{\textless $\textit{img}_{#1}$\textgreater{}}  
\newcommand{\imgTokenEnd}[1]{\textless /$\textit{img}_{#1}$\textgreater{}}  
\newcommand{\phraseToken}{\textless \textit{caption}\textgreater{}}  
\newcommand{\phraseTokenEnd}{\textless /\textit{caption}\textgreater{}}  
\newcommand{\imgContent}{\{\textit{img\_content}\}}  
\newcommand{\hashTriple}{\#\#\#}

\newcommand{\groundingToken}{\textless \textit{grounding}\textgreater{}}
\newcommand{\patchIndex}[1]{\textless $\textit{loc}_{#1}$\textgreater{}}  
  
\newcommand{\flame}{\includegraphics[width=0.4cm]{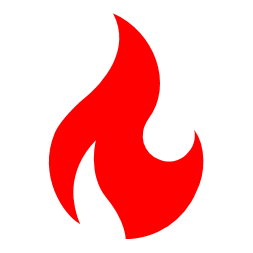}}  
\newcommand{\snowflake}{\includegraphics[width=0.4cm]{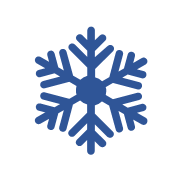}}  
\usepackage{wrapfig,lipsum,booktabs}

\usepackage{longtable}

\usepackage{enumitem}
\usepackage{bm}

\newcommand{\ieno}{\textit{i}.\textit{e}.}
\newcommand{\egno}{\textit{e}.\textit{g}.}
\newcommand{\etcno}{\textit{etc.}}
\newcommand{\ours}{RelationVLM}
\newcommand{\short}{LVLMs}
\newcommand{\shortno}{LVLM}

\begin{document}

\title{RelationVLM: Making Large Vision-Language Models Understand Visual Relations}

\titlerunning{RelationVLM}

\author{
Zhipeng Huang\textsuperscript{\rm 1,2}\thanks{Equal Contributions.}\thanks{This work was done when Zhipeng Huang was an intern at MSRA.} \quad 
Zhizheng Zhang\textsuperscript{\rm 2*}\thanks{Corresponding Author.} \quad
Zheng-Jun Zha\textsuperscript{\rm 1}\quad \\
Yan Lu\textsuperscript{\rm 2}\quad
Baining Guo\textsuperscript{\rm 2}\\
}
\authorrunning{Zhipeng et al.}

\institute{\textsuperscript{\rm 1}University of Science and Technology of China \quad 
\textsuperscript{\rm 2}Microsoft Research Asia \\
\email{hzp1104@mail.ustc.edu.cn} \quad \email{zhazj@ustc.edu.cn} \\
\email{zhizzhangms@gmail.com} \quad
\email{\{yanlu,bainguo\}@microsoft.com}
}

\maketitle

\begin{abstract}

The development of Large Vision-Language Models (\short) is striving to catch up with the success of Large Language Models (LLMs), yet it faces more challenges to be resolved. Very recent works enable \short~to localize object-level visual contents and ground text to them. Nonetheless, current \short~still struggle to precisely understand visual relations due to the lack of relevant data. In this work, we present RelationVLM, a large vision-language model capable of comprehending various levels and types of relations whether across multiple images or within a video. Specifically, we devise a multi-stage relation-aware training scheme and a series of corresponding data configuration strategies to bestow RelationVLM with the capabilities of understanding semantic relations, temporal associations and geometric transforms. Extensive case studies and quantitative evaluations show RelationVLM has strong capability in understanding such relations and emerges impressive in-context capability of reasoning from few-shot examples by comparison. This work fosters the advancements of \short~by enabling them to support a wider range of downstream applications toward artificial general intelligence.
\end{abstract}

\section{Introduction}
\vspace{-1mm}
The success of Large Language Models (LLMs)~\cite{brown2020language,openai2023gpt4,touvron2023llama} has advanced the development of large multimodal models, especially for Large Vision-Language Models (\short)~\cite{alayrac2022flamingo,li2023blip,liu2023visual}. Recent studies~\cite{zhang2023meta,wu2023next} enable the capabilities of understanding and processing the information from other modalities based on pre-trained LLMs, expanding their application ranges and opening countless possibilities in developing general-purpose visual understanding system.

\begin{figure*}[!t]
	\centering
	\includegraphics[width=\textwidth]{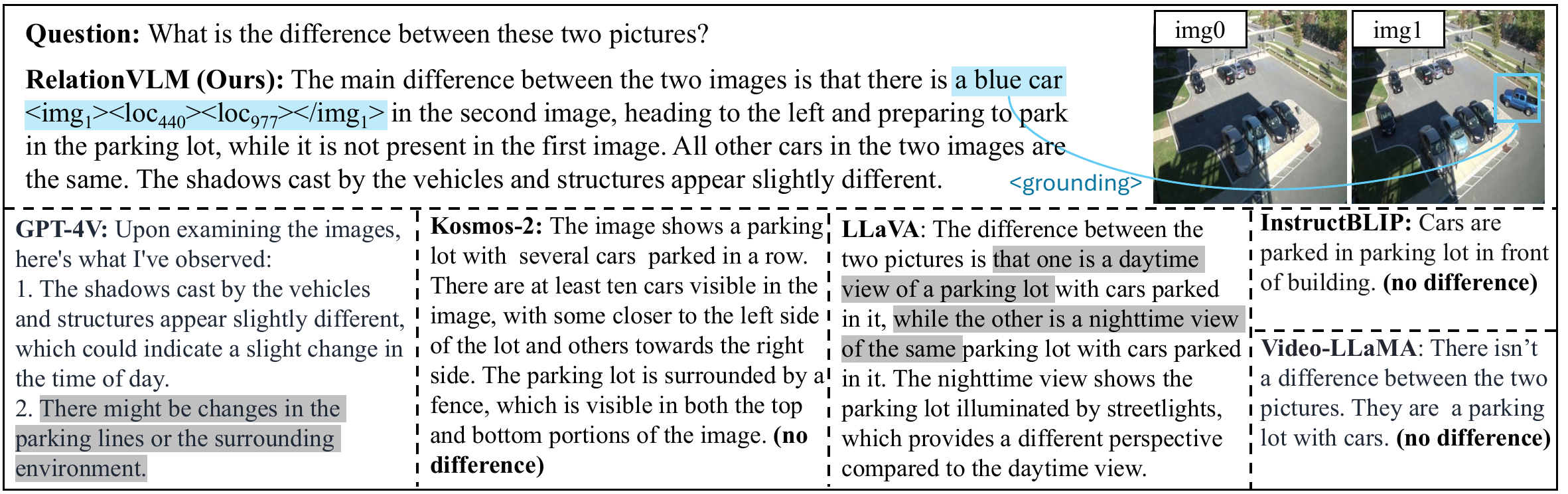}
    \vspace{-4mm}
	\caption{In a simple two-image comparison task, our~\ours~highlights differences using grounded bounding boxes, while other methods miss correct differences (gray highlight) or claim no differences (marked with bold 'no difference'). `\imgToken{1}' and `\imgTokenEnd{1}' denote the second image, while `\patchIndex{440}\patchIndex{977}' refers to the bounding box's left-top and right-bottom coordinates. More details see Sec.~\ref{sec:data_construction}.}
	\label{fig:abstract}
    \vspace{-6mm}
\end{figure*}

Very recent research efforts on \short~\cite{liu2023visual,instructblip,zhu2023minigpt,peng2023kosmos} have been dedicated to enabling \short~to grasp correspondence between different modalities, exhibiting impressive capabilities on captioning and grounding. Nevertheless, compared to the enterprise-level performance of LLMs, the capabilities of understanding and processing multimodal signals by \short~are still in their early stages. Besides the correspondence between visual features and language, we also expect \short~to precisely understand the relations within the visual features themselves, thereby learning to make visual comparison as humans. This is, in fact, very critical for many downstream tasks in a wider range, \egno, visual retrieval, anomaly detection, video understanding. By empirically evaluating the existing \short~as shown in Figure \ref{fig:abstract}, including video-based \short~\cite{zhang2023video}, we find they all struggle to precisely perceive and understand visual relations across different images or frames. Detailed comparison results are described in the caption of Figure \ref{fig:abstract}.

Visual relations have a rich variety of types in nature, mainly including semantic relations (\ieno, whether visual objects have the same semantics), temporal associations (\ieno, the order of events in time), and geometric transforms (\ieno, spatial deformations). Building a \shortno~capable of comprehensively understanding them (abbreviated as relation-aware \shortno~) is actually a challenging thing. This is because understanding these diverse visual relations requires handling data of higher dimensions than those handled by LLMs. Such capability is also hard to emerge directly after training with straightforwardly interleaved image-text pair data, as shown by the failure cases in Figure \ref{fig:abstract}. Training such a \shortno~from scratch requires an abundance of annotated image-text pairs, thus being notably costly. In this paper, we propose an efficient method for enabling \short~to understand visual relations with a pre-trained vision encoder and a pre-trained LLM as the language decoder. Moreover, we utilize the off-the-shelf annotations from the existing datasets with our proposed data configuration strategies, obviating the need for extra annotations about relations.

We introduce~\ours, a large vision-language model that not only has the grounding capability but also comprehends various visual relations.~\ours~is built in the lowest costs as possible, by making full use of the knowledge already mastered by pre-trained models.~\ours~employs a pre-trained vision encoder, a pre-trained LLM and an adapter implemented by a linear layer, where we solely need to train the adapter and fine-tune the LLM with LoRA~\cite{hu2022lora} for reaching the desired goal. To achieve this, a three-stage training strategy is devised. In the first stage, we adopt coarsely interleaved image-text pairs for feature alignment across modalities. In the second stage, we extract relation-related attributes from the existing public datasets upon their available annotations, and automatically re-organize them into dialogue-form data using GPT-4~\cite{openai2023gpt4}.~\ours~grasp primary capabilities on understanding various relations through generative training with these data in this stage. Details on data creation and configuration are introduced in the method part. In the third stage, we perform instruction tuning with a combined dataset consisting of existing visual instruction tuning datasets and a high-quality subset manually selected from the dataset generated in the second stage.

After these three stages of training, our~\ours~is not only able to accurately understand various relations across multiple images or within a video, but also demonstrates impressive in-context learning capabilities in unseen visual comparison application tasks, such as medical diagnosis and anomaly detection. By providing just a few examples in prompts, our model can robustly generalize its visual comparison capabilities and accurately apply them to specific real-world domains even though they are unseen before.

In summary, we make the following contributions in this work:

\begin{itemize}[leftmargin=*]
\item We build~\ours, which addresses the shortcomings of current Large Vision-Language Models (\short) in their inability to accurately comprehend various visual relations, including semantic relations, temporal associations and geometric transforms. It endows \short~with general-purpose visual comparison capabilities, being a step forward towards achieving general-purpose visual understanding system.

\item We provide a data construction scheme for extracting relation attributes from existing public datasets and adopt a LLM (GPT-4) to automatically organize them into an appropriate form for multimodal generative training, for enabling \short~to comprehend various visual relations.

\item We qualitatively and quantitatively evaluate our built~\ours~in comprehending different types of relations. Besides, we also showcase the visual in-context learning and generalization of our~\ours~for unseen real-world visual comparison tasks, \egno, medical diagnosis and anomaly detection.

\end{itemize}

\section{Related Work}

\label{sec:related_work}

\vspace{-1mm}
\subsection{Image-based \short}
\vspace{-1mm}
Existing approaches on image-based~\short~can be categorized into integrated vision-language systems and end-to-end \short. The former, such as Visual ChatGPT~\cite{wu2023visual}, MM-REACT~\cite{yang2023mm},  HuggingGPT~\cite{shen2023hugginggpt}, and InternGPT~\cite{liu2023internchat}, integrate various existing vision models or tools into a centralized LLM controller with neural language prompts, without trainable parameters, excelling at well-defined problems but potentially lacking zero-shot abilities for open-ended instructions. The latter, including Flamingo~\cite{alayrac2022flamingo}, BLIP2~\cite{li2023blip}, MiniGPT4~\cite{zhu2023minigpt}, LLaVA~\cite{zhang2023transfer}, mPLUG-Owl~\cite{ye2023mplugowl}, PaLM-E~\cite{driess2023palme}, KOSMOS~\cite{huang2023language}, employ vision encoders and finetune adapter layers for cross-modality embedding alignment. And they either freeze LLM, finetune with LoRA~\cite{hu2022lora} or end-to-end fine-tune LLM. Our work belongs to the latter,~\ieno, end-to-end~\short, but we introduce a novel training approach that significantly improves the model's capacity to comprehend relationships among multiple images, beyond other end-to-end \short. 

\vspace{-1mm}
\subsection{Video-based \short}
\vspace{-1mm}
Video-based~\short~focus on processing sequential video frames as the model inputs. Video-Chat~\cite{li2023videochat} and Video-LLaMA~\cite{zhang2023video} integrate video foundation models with LLMs along with cross-modality adapters. Video-ChatGPT~\cite{maaz2023video} introduces a novel annotation framework and proposes a quantitative video evaluation framework. Although, video-based~\short~support multiple images as input, they exhibit certain limitations on understanding visual relations. These models use frames derived from the same video, resulting in minimal scene variation and negligible changes between images. Consequently, relationships between images are predominantly limited to simple spatial motion. Additionally, the captions used for training often provide coarse descriptions of temporal actions only, lacking the fine-grained details necessary for accurately depicting the relationships between frames. In contrast, our work addresses not only temporal associations but also incorporates fine-grained semantic relations and geometric transforms which enables our model to capture more comprehensive and detailed relational information, surpassing the limitations of existing video-based~\short~on understanding visual relations.


\vspace{-2mm}
\subsection{Visually-grounded \short}
\vspace{-1mm}
Visually-grounded \short~\cite{chen2023shikra,peng2023kosmos,chen2023minigptv2,wang2023cogvlm,zhao2023svit,zhang2023llavagrounding} are a subset of Image-based \short, distinguished by their use of grounding data during training. Leveraging the grounding training data, Visually-grounded \short~excel at not only generating textual descriptions from visual inputs but also at localizing and identifying specific visual elements that correspond to nouns or entities within the generated text. However, due to their training data being formatted for single-image input, they lack the ability to concurrently process multiple images, let alone analyze relationships between different images. Our work extends them by introducing an innovative training methodology that substantially enhances the model's ability to understand and interpret relationships among multiple images.

\begin{figure*}[!t]
	\centering
	\includegraphics[width=\textwidth]{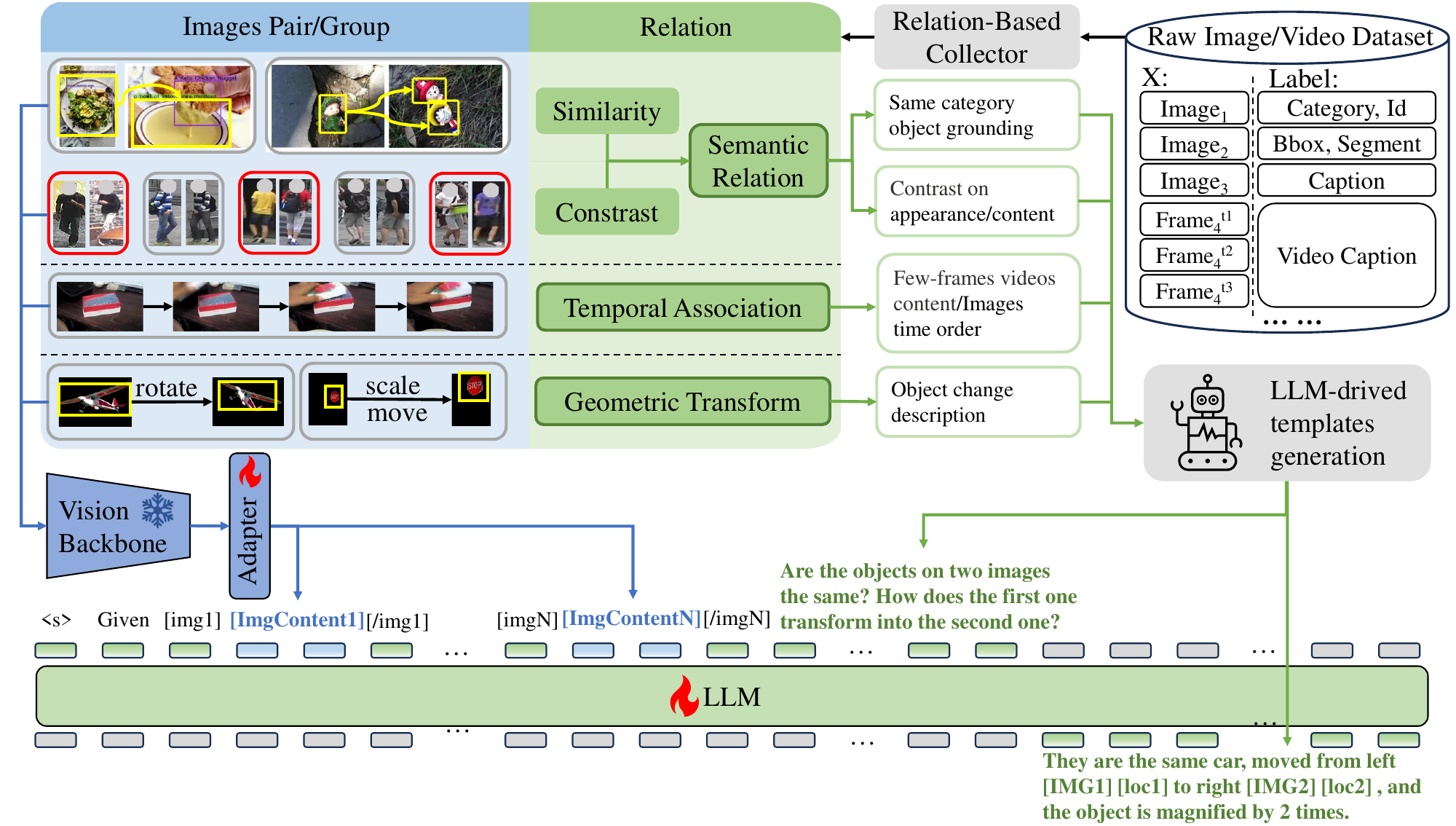}
	\caption{The overall framework of~\ours. The \ours~consists of a frozen vision encoder, a learnable adapter and a learnable LLM. We employ publicly available datasets and adopt GPT-4 to curate them into a dialogue form for its training.}
    \vspace{-6mm}
	\label{fig:framework}
\end{figure*}


\section{\ours}
\vspace{-1mm}

\subsection{Definition of Visual Relations}
\label{sec:definition}
\vspace{-1mm}

We aim to build a \shortno~that can understand various visual relations. As an initial study, we start with the three most basic and widely applied high-level categories of visual relations upon their intrinsic characteristics: semantic relations (similarity/contrast), temporal associations and geometric transforms. Semantic relations cover similarity (categorically alike objects should be spatially grounded) and contrast (differences in category and attributes). Temporal associations address movements over time, while geometric transformations consider spatial changes. A certain type of visual relation may be manifested at different attributes. For example, the semantic relations of two objects may reside in their categories, colors, shapes, \etc. Future research will explore additional low-level visual relations \egno, corruption, lighting conditions. In this work, we extract relation annotations from existing public datasets including diverse semantic labels for categories, attributes, bounding boxes, \etcno, and automatically organize them into a dialogue form for generative training.

To clearly describe our targeted relations and the construction process of their labels, we formulate a specific subdivided relation as a function denoted by $R_n$ where $n$ indexes the relation functions. Given a dataset $\mathcal{D} = \{(x_i, y_i)\}_{i=0}^{N}$ that includes $N$ images and  $N_R$ types of relations in its annotations, we have a set of relation functions $\mathcal{R} = \{R_n\}_{n=0}^{N_R}$. Here, $x_i$ and $y_i$ represent a sample and its corresponding semantic label, respectively. $R_n$ maps the semantic labels of two given samples into a binary value, wherein $R_n(y_i, y_j) = 1$ denotes there exists the relation corresponding to $R_n$ between $x_i$ and $x_j$, and $R_n(y_i, y_j) = 0$ indicates there is no such relation. This definition is also applicable to image groups involving more than two images.

\vspace{-1mm}
\subsection{Data Construction}
\label{sec:data_construction}
\vspace{-1mm}

\noindent\textbf{General introduction of the data construction process.} We introduce data construction for training \ours, which is a dominant aspect for enabling \short~to understand various visual relations. In light of the enormous costs of annotating visual relations at scale, we extract the required labeled data from existing public datasets purposefully to cover different types of visual relations introduced in Sec.\ref{sec:definition}. For semantic relations, we utilize the datasets containing reference expressions and spatial localization information for objects and persons, including GRIT~\cite{peng2023kosmos}, refCOCO~\cite{yu2016modeling}, person reid datasets~\cite{zheng2015scalable,li2014deepreid,xiao2017joint,wei2018person,li2015deepmar,8510891}, CUB-200-2011~\cite{wah2011caltech} and MIMIC~\cite{li2023mimicit}. These datasets inherently carry explicit semantic annotations for entities, enabling the effortless acquisition of labels for semantic relationships between two entities based on the consistency of their labels, such as whether they belong to the same category or not. For temporal associations, we adopt the video datasets with the captions for activities and the frame IDs that can describe the chronological orders. Specifically, we use SSv2~\cite{goyal2017something} and WebVid~\cite{bain2021frozen}. Regarding geometric transform, we cannot find off-the-shelf annotations for using. Thus, we segment natural images and perform geometric transformations (including Horizontal flip, vertical flip, brightness adjustment, rotation, scaling, and moving) on the segmented objects for synthesizing the needed dataset. Although the dataset is synthesized, we strive to maintain a broad diversity of the synthesized data to enhance the generalization capability of the model as much as possible. Datasets for learning different types of visual relations are jointly used.

In Sec.\ref{sec:definition}, given a dataset with off-the-shelf annotations, we formulate what image pairs or image groups can be considered to have a certain visual relation. To construct a new dataset containing diverse visual relations, we first collect the image pairs or groups with one or more defined visual relations from the aforementioned datasets. We encode the original annotations provided in the raw datasets into a linguistic form, and then utilize a mature LLM (GPT-4) to automatically generate natural language descriptions for visual relations via prompt engineering. Subsequently, we further adopt GPT-4 to further convert linguistic relation descriptions into a dialogue (\ieno, question-answering) form for generative training. Mathematically, taking an image pair $(x_i, x_j)$ as an example, such data construction process can be formulated as:

\vspace{-5mm}
\begin{equation} \label{eq:1}
    y^{dialog}_{i,j} = LLM^{dialog}(LLM^{desc}(E(y_i^{raw}), E(y_j^{raw}), p^{desc} | R_n), p^{dialog} | R_n),
\end{equation}
\vspace{-5mm}

where $y_i^{raw}$ and $y_j^{raw}$ are the raw annotations of $x_i$ and $x_j$, respectively. They satisfy the condition $R_n(y_i^{raw}, y_j^{raw})=1$, indicating that $x_i$ and $x_j$ have the relation corresponding to the relation function $R_n$. The $E(\cdot)$ represents the rule-based data processing function of encoding the raw annotations from their original form to be a linguistic form, $LLM^{desc}(\cdot)$ denotes the function of generating natural language captions with relation-related descriptions involved via a mature LLM, and $LLM^{dialog}(\cdot)$ denotes the function of converting natural language caption into a dialog form with a LLM. $p^{dialog}$ and $p^{desc}$ are the prompts for LLMs. They are manually designed and condition on the type of relation, \ieno, $R_n$. $p^{desc}$ is shown in Figure~\ref{fig:prompts1} and $p^{dialog}$ is shown in the supplementary.

\begin{figure*}[!t]
	\centering
	\includegraphics[width=0.8\textwidth]{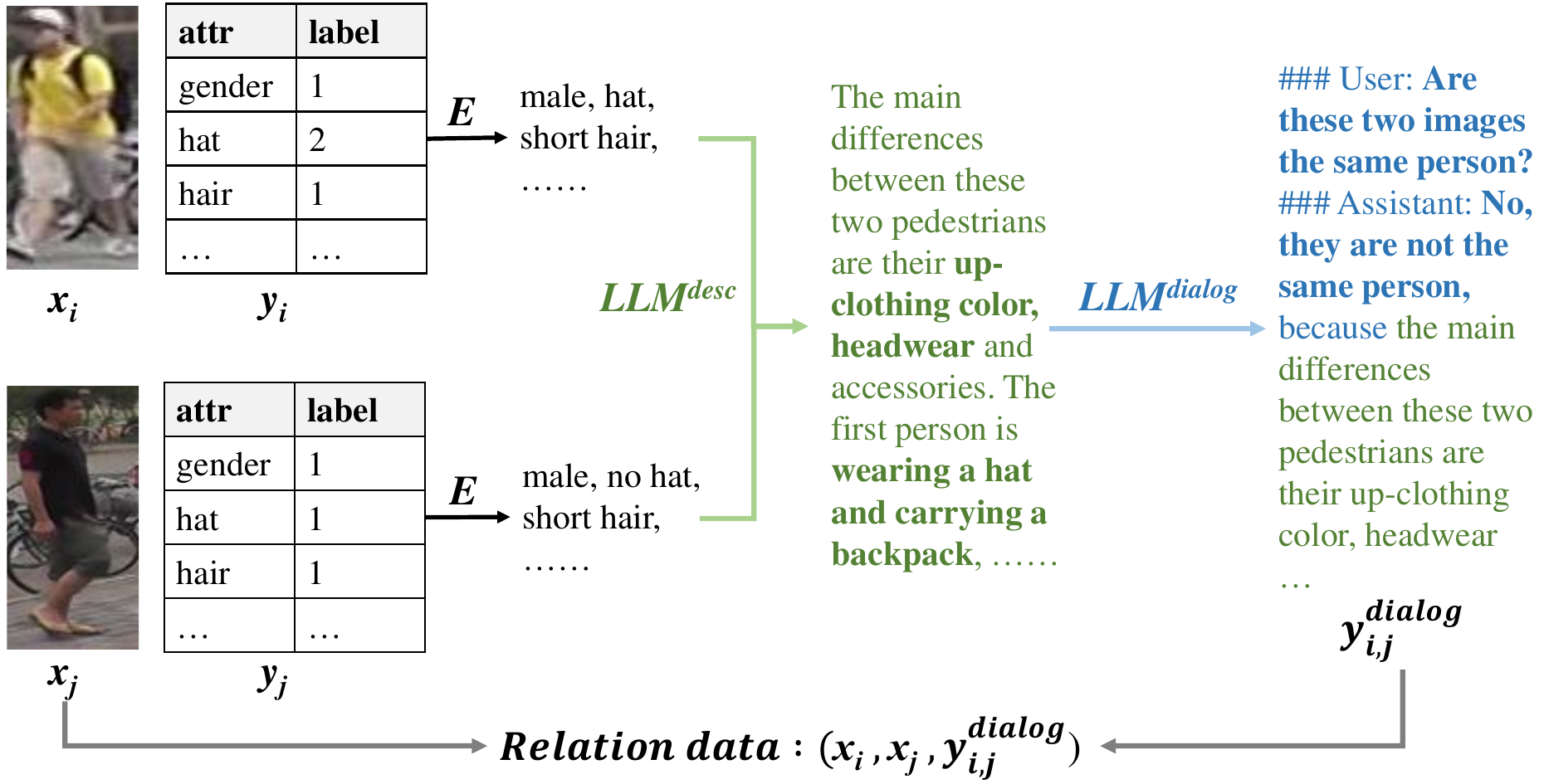}
	\caption{Illustration of the data construction process.}
	\label{fig:data_construction}
 \vspace{-6mm}
\end{figure*}

\noindent\textbf{Representative instantiation of data construction.} We generally introduce the data construction process as above. Here, we describe a representative example for further clarification. As shown in Figure~\ref{fig:data_construction}, given two images $(x_i, x_j)$ in RAP~\cite{li2015deepmar}, each containing a unique individual, their raw annotations $(y^{raw}_i, y^{raw}_j)$ provides the information of person IDs and a series of attributes. Let $R_{n1}(\cdot)$ be a relation function corresponding to the relation of different IDs in the sense that $R_{n1}(y^{raw}_i, y^{raw}_j)=1$ indicates $x_i$ and $x_j$ are two different persons. Besides $R_{n1}(\cdot)$, we have other relation functions $R_n(\cdot)$ with different values of $n$ for describing other types of relations, such as the ones for clothing.

We first transform their raw annotations into a linguistic form individually using a rule-based data processing function, getting the results as follows:

\vspace{-2mm}
\begin{tcolorbox}[breakable]
\vspace{-1mm}
\textbf{$E(y_i^{raw})$}: person, ID: 128, male, short hair, short sleeve, short lower body clothing, pants, no hat, no backpack, no bag, no handbag, teenager, wearing a gray upper-body clothing, wearing black lower-body clothing. \\
\textbf{$E(y_j^{raw})$}: person, ID: 334, male, short hair, short sleeve, short lower body clothing, pants, no hat, backpack, no bag, no handbag, teenager, wearing a black upper-body clothing, wearing black lower-body clothing, backpack bbox on \patchIndex{256} \patchIndex{489}.
\vspace{-1mm}
\end{tcolorbox}
\vspace{-2mm}

The location tokens \patchIndex{256} and \patchIndex{489} are the encoded locations of the top-left and bottom-right points of the bounding boxes, respectively. The encoding method for location tokens is from the commonly used one introduced in~\cite{peng2023kosmos,chen2021pix2seq}, which represents the positions in an image by patchifying it into discretized bins.

Then, we prompt GPT-4 to automatically generate linguistic relation description for $(x_i, x_j)$ based on the relation function $R_n(\cdot)$ they satisfy. As formulated in Eq. (\ref{eq:1}), this process corresponds to $y_{i,j}^{desc}\!=\!LLM^{desc}(E(y_i^{raw}), E(y_j^{raw}), p^{desc} | R_n)$. The used prompts $p^{desc}$ are detailed in Figure~\ref{fig:prompts1}, and the obtained results are as below:

\vspace{-2mm}
\begin{tcolorbox}[breakable]
\vspace{-1mm}
\textbf{$y_{i,j}^{desc}$:} The main difference between them is the color of their upper-body clothing and the presence of a backpack. The first person is wearing a gray shirt while the second is wearing a black shirt. Additionally, the second person is carrying \phraseToken a backpack \phraseTokenEnd \imgToken{1} \patchIndex{256} \patchIndex{489} \imgTokenEnd{1}, while the first person is not carrying any bags or backpacks.
\vspace{-1mm}
\end{tcolorbox}
\vspace{-2mm}

In the illustrated $y_{i,j}^{desc}$ above, \phraseToken and \phraseTokenEnd are special tokens that indicate the beginning and the end of available semantic descriptions for the attachments. \imgToken{1} and \imgTokenEnd{1} tell which image the attachment belongs to. \patchIndex{256} and \patchIndex{489} indicate the location of the bounding box.

Furthermore, we convert $y_{i,j}^{desc}$ into a dialog (question-answering) form for generative training as Eq. \ref{eq:1}. This step can enrich the produced data so as to improve the robustness and alignment of \ours. Detailed implementations of the used $p^{dialog} | R_n$ are provided in the supplementary. One of representative template for the finally generated $y^{dialog}_{i,j}$ is as below:

\vspace{-1mm}
\begin{tcolorbox}[breakable]
\vspace{-1mm}
\textbf{$y_{i,j}^{dialog}$:} \hashTriple~Human: \imgToken{0} \imgContent \imgTokenEnd{0}, \imgToken{1} \imgContent \imgTokenEnd{1} \
Are the two people in the two images the same person? \groundingToken \\
\hashTriple~Assistant: No, they are not the same person. \{$y_{i,j}^{desc}$\} Therefore, based on these differences, it can be concluded that these two people are not the same person.
\vspace{-1mm}
\end{tcolorbox}
\vspace{-1mm}

Here, \imgToken{0} \imgTokenEnd{0} and \imgToken{1} \imgTokenEnd{1} points to different images. \imgContent~is a placeholder, which will be replaced by the corresponding image tokens in the embeding space. The \groundingToken~is a special token that informs \ours~that the subsequent answer should include grounding bounding boxes explicitly. As of now, one sample ${(x_i, x_j), y_{i,j}^{dialog}}$ for training \ours~is produced.

\subsection{Model Training}
As shown in Figure~\ref{fig:framework}, \ours~consists of a vision encoder, an adapter and a LLM-based decoder, following the commonly used architecture design for \short~\cite{li2023blip,zhang2023transfer,zhu2023minigpt,ye2023mplugowl}. The vision encoder, composed of a VIT-based vision backbone and a Q-Former, encodes images into a set of visual tokens. The adapter achieves cross-modality alignment, which is implemented by a linear layer. The decoder could be a pre-trained LLM where we use Vicuna~\cite{vicuna2023} in this paper. For its training, a relation-based collector collects image pairs or groups from raw labeled datasets. Their original labels are input into the LLM using prompts based on the relation type. The LLM generates fine-grained natural language descriptions along with referring bounding box. The image pairs/groups are then input into a frozen vision backbone to obtain vision tokens, which are transformed through an adapter and combined with language tokens for training the LLM using LoRA. We perform the typical generative training for \ours~by maximizing the likelihood of each expected token based on preceding tokens over the decoding sequence.

\section{Experiments}
\label{sec:experiment}

\vspace{-2mm}

\subsection{Implementation}

In this section, we present critical details of data construction and model training, with more implementation details provided in the supplementary.

\begin{figure*}[!t]
	\centering
    \vspace{-2mm}
	\includegraphics[width=\textwidth]{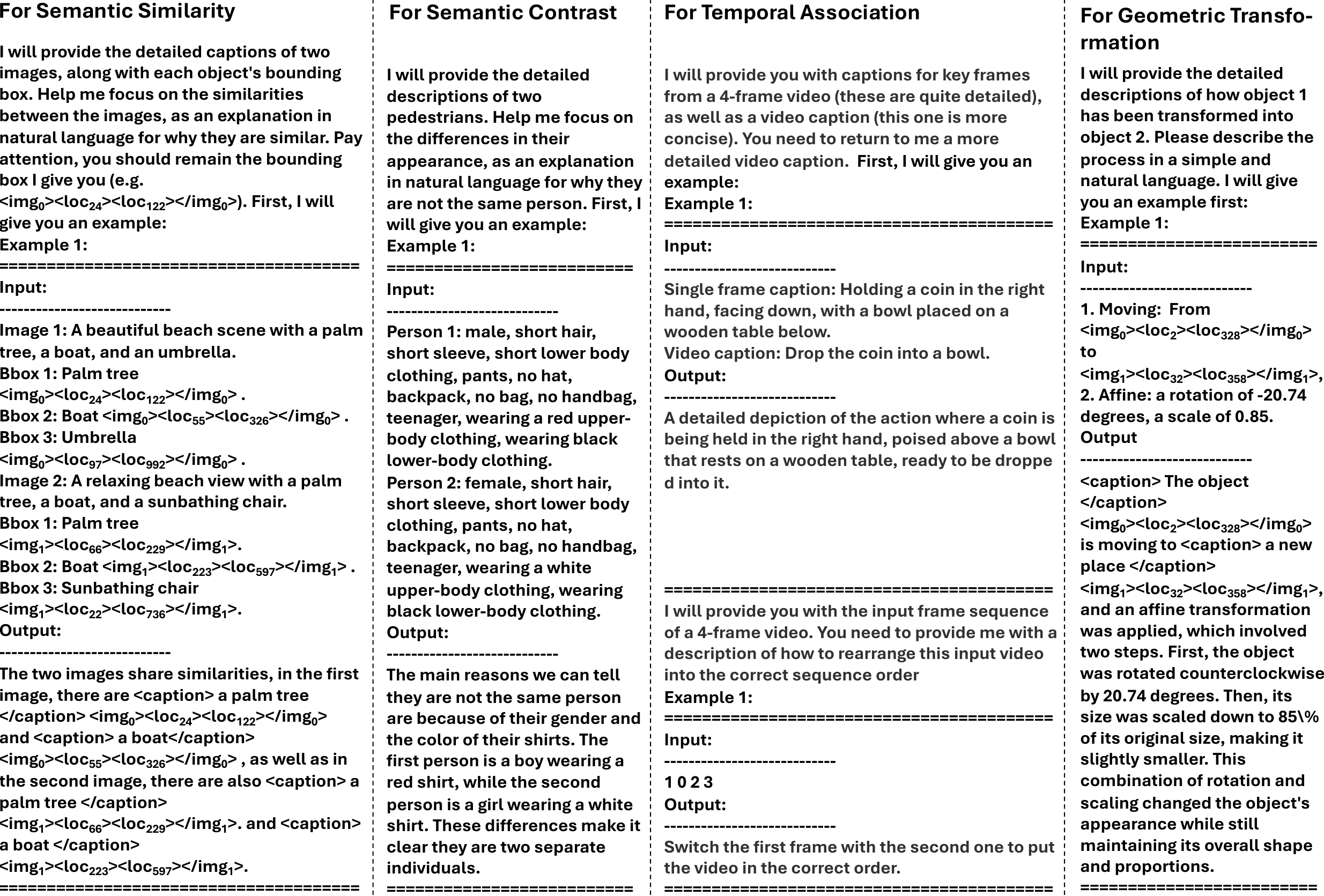}
    \vspace{-4mm}
	\caption{Illustrative prompts for identifying and extracting the investigated visual relation from annotated datasets, corresponding to  $p^{desc}$ in Eq.~\ref{eq:1}.}
    \vspace{-4mm}
	\label{fig:prompts1}
\end{figure*}

\noindent\textbf{Data construction prompts.} As introduced in Sec.~\ref{sec:data_construction}, to construct relation-contained data, after encoding the raw annotations from their original form to be a linguistic form, we prompt GPT-4 to automatically identify and extract linguistic relations to avoid substantial manual annotation costs.
The prompts are illustrated in Figure~\ref{fig:prompts1} which are categorized into four segments, each corresponding to the types of relations we investigate. Specifically, for the similarity semantic relation, GPT-4 is prompted to identify similarities between images based on identical nouns or the same category of nouns. It provides a description of the similarities between two images, marking the location of these similar nouns within each image. For the contrast semantic relation, GPT-4 is prompted to detect differences in image content. If the original data contains object location label, it is retained for grounding. For temporal association, prompts are geared towards describing content within frames of a video or reconstructing the sequence of shuffled frames for building data that challenges the model’s understanding of temporal association. Lastly, the geometric relation is based on random geometric transformations, generating corresponding linguistic descriptions that reflect these changes. 
Additional detailed prompts, $p^{dialog}$ and evaluation prompts are provided in the supplementary.

\noindent\textbf{Training details.} The vision backbone comprises 1B parameters and is initialized using BLIP-2~\cite{li2023blip} pretrained weights. The employed LLM model has 7B parameters, initialized with Vicuna-v0~\cite{vicuna2023} weights. And the adaptor is trained from scratch. Training was conducted on 4$\times$A100(80G) GPUs. We leverage the Zero-2 optimization, facilitated by the DeepSpeed framework \cite{rajbhandari2020zero, rasley2020deepspeed}. The entire training process spanned 5 days and was divided into 3 stages. Detailed descriptions of our phased training strategy, configuration and the datasets utilized for each stage are provided in the supplementary.

\vspace{-3.5mm}
\subsection{Qualitative Results}
\label{sec:qualitative}

\begin{figure*}[!t]
	\centering
	\includegraphics[width=\textwidth]{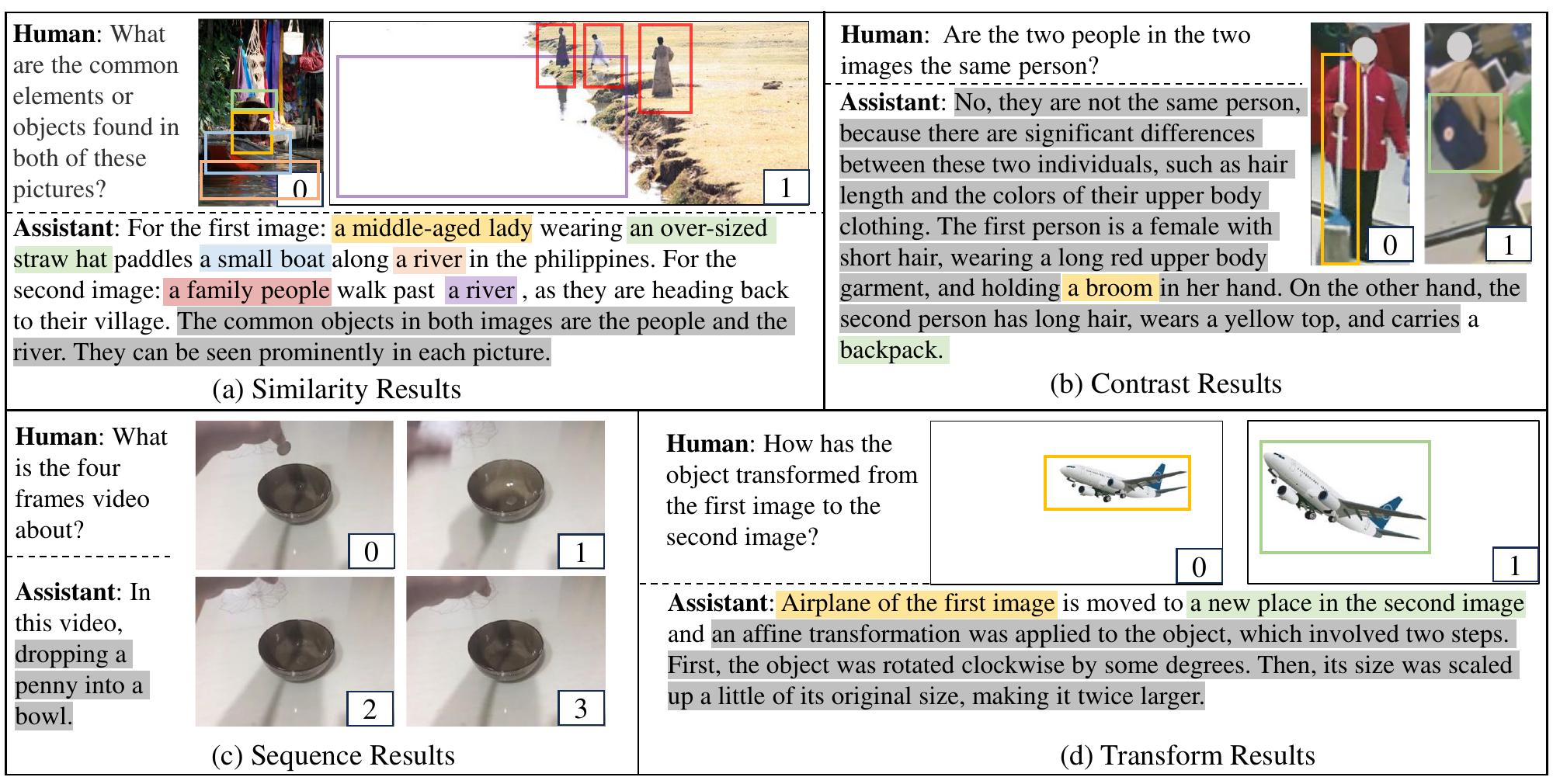}
    \vspace{-2mm}
	\caption{Examples of~\ours~answering questions about image relations, including Semantic Relation (a)(b), Temporal Association (c), and Geometric Transformation (d). Color-highlighted text are output along with referring the same color bounding box. And the grey-highlighted text shows details relation description provided by~\ours.}
	\label{fig:main_result}
    \vspace{-4mm}
\end{figure*}

We show the qualitative results by providing representative case study results in Figure \ref{fig:main_result}. As the example (a) shows, \ours~can not only perform grounding within a single image like Kosmos-2~\cite{peng2023kosmos}, but also identify the contents of different images and accurately perceive the commonality between the two images. The example (b) presents that \ours~can distinguish between two different persons and further enumerate the differences in their attributes as the basis for judgment. The example (c) shows that \ours~is able to perceive the temporal association so as to recognize the happened activity in a video. The example (d) exhibits that \ours~is capable of accurately analyzing the geometric transformations that have occurred to the same object in the two images. These results explicitly demonstrate that our \ours~can understand different types of visual relations across multiple images or within a video. Additional detailed qualitative results that describe the relationships across multiple images or within a single image are in the supplementary.

\subsection{Quantitative Results}
\label{sec:quantitative}
\vspace{-2mm}



\noindent\textbf{Evaluation Metric.} From a qualitative perspective, we explicitly showcase the capability of \ours~in comprehending various visual relations. Besides, we also perform quantitative evaluation with the following metrics:

1) Traditional metrics: We extract task-related results from the outputs of \ours, and use traditional metrics to assess their accuracies. For classification, we report the accuracy (Acc), precision (Prec), recall (Rec), and F1-score (F1). For grounding, following~\cite{yu2016modeling,liu2023grounding,chen2023shikra,peng2023kosmos}, we calculate the accuracy of predicted bounding boxes, named ``BBox Acc''. The predicted bounding box is considered correct when its IoU is greater than 0.5, and incorrect otherwise. It is considered as incorrect as well when it cannot be decoded successfully. 

2) LLM-based metric:  Traditional metrics can only measure partial information of the model outputs. We employ a mature LLM (GPT-4~\cite{openai2023gpt4}) to assess the overall plausibility of RelationVLM's outputs. We name this metric Relation Score (RS), and detail the evaluation prompts in the supplementary. The range of RS is in [1-5] (the bigger RS, the better). We admit that assessing the linguistic quality produced by Large Language Models quantitatively continues to be a challenge that has not been fully addressed. Though leveraging GPT-4 to serve as an evaluator is a prevalent solution~\cite{zhang2023transfer,Maaz2023VideoChatGPT} so far, we also include human side-by-side evaluation in the supplementary.

\noindent\textbf{Settings.} Considering the inference efficiency of LLMs, we sample some subsets from existing public datasets to evaluate the capabilities of \ours~in different aspects. Specifically, we use the subset sampled from the test set of COCO~\cite{lin2014microsoft} to evaluate the capabilities of \ours~in identifying the common features among multiple images and referring expression comprehension (\ieno, grounding). Moreover, we construct a test set from the test sets of~\cite{xiao2017joint,wei2018person,li2014deepreid,zheng2015scalable} to evaluate the capabilities of \ours~in differentiating multiple similar images and identifying their detailed differences. We construct a test set from Something-Something-v2 (SSv2~\cite{goyal2017something}) and ActivityNet~\cite{caba2015activitynet} to assess the capability in understanding the temporal associations within videos, and construct another one from COCO~\cite{lin2014microsoft} to evaluate the capability in understanding geometric transforms. More details about the datasets used for quantitative evaluation can be found in Supplementary.

\begin{table}[!t]
  \centering
    \vspace{-2mm}
    \caption{Comparative comparison of \ours~with the state-of-the-art image-based, video-based, and visually-grounded \short, across semantic similarity, semantic contrast, temporal association and geometric transformation tasks. `SSv2' denotes Something-Something-v2 dataset. `AN' represents activityNet Dataset. `RS' denotes Relation Score.}
    \vspace{-2mm}
    \resizebox{\linewidth}{!}{
    \begin{tabular}{l|cc|cc|cc|cc}
    \toprule
    \multicolumn{1}{c|}{\multirow{2}[0]{*}{Models}} & \multicolumn{2}{c|}{Similarity} & \multicolumn{2}{c|}{Contrast} & \multicolumn{2}{c|}{Temporal} & \multicolumn{2}{c}{Geometric} \\
          & BBox Acc(\%) & RS    & Acc (\%) & RS    & RS@SSv2 & RS@AN & BBox Acc(\%) & RS \\
    \midrule
    LLaVA~\cite{zhang2023transfer} & - & 1.4   & 58.2  & 1.6   & 0.3   & 1.2   & - & 0.6 \\
    MM-GPT~\cite{gong2023multimodal} & - & 1.2   & 53.2  & 1.1   & 0.3   & 1.1   & - & 0.5 \\
    mPLUG-Owl~\cite{ye2023mplugowl} & - & 1.2   & 56.2  & 1.5   & 0.9   & 1.6   & - & 0.5 \\
    Open-Flamingo~\cite{awadalla2023openflamingo} & - & 1.7   & 50.2  & 1.6   & 0.6   & 1.2   & - & 0.8 \\
    Otter~\cite{li2023otter} & - & 1.9   & 51.7  & 1.6   & 0.6   & 1.7   & - & 0.9 \\
    \hline
    Video-Chat~\cite{li2023videochat} & - & 0.7   & 50.9  & 1.3   & 1.7   & 2.2   & - & 1.2 \\
    Video-LLaMA~\cite{zhang2023video} & - & 0.9   & 52.1  & 1.5   & 0.9   & 2.0   & - & 1.6 \\
    Video-ChatGPT~\cite{Maaz2023VideoChatGPT} & - & 1.1   & 54.3  & 1.6   & 1.9   & 2.4   & - & 1.9 \\
    \hline
    Shikra~\cite{chen2023shikra} & 42.8  & 1.7   & 58.7  & 1.9   & 0.5   & 1.1   & 51.7  & 1.0 \\
    Kosmos-2~\cite{peng2023kosmos} & 41.3  & 1.6   & 60.2  & 1.8   & 0.5   & 1.2   & 55.3  & 1.1 \\
    MiniGPT4-v2~\cite{chen2023minigptv2} & 38.7  & 1.8   & 52.2  & 1.5   & 0.6   & 1.3   & 47.3  & 0.8 \\
    CogVLM~\cite{wang2023cogvlm} & 44.5  & 2.1   & 63.3  & 1.9   & 0.8   & 1.5   & 59.7  & 1.8 \\
    SVIT~\cite{zhao2023svit}  & 41.1  & 1.6   & 58.6  & 1.5   & 0.8   & 0.9   & 56.3  & 1.3 \\
    LLaVA-Grounding~\cite{zhang2023llavagrounding} & 41.8  & 1.8   & 59.8  & 1.8   & 0.9   & 0.9   & 57.0  & 1.4 \\
    \hline
    Baseline~\cite{zhu2023minigpt} & - & 1.7   & 51.3  & 1.6   & 0.7   & 1.4   & - & 0.8 \\
    \rowcolor[gray]{0.9}\ours & \textbf{49.3}  & \textbf{2.5}   & \textbf{83.2}  & \textbf{3.3}   & \textbf{3.2}   & \textbf{2.4}   & \textbf{69.7}  & \textbf{3.2}  \\
    \bottomrule
    \end{tabular}%
    }
  \label{tab:compare_lvlm_relation}%
  \vspace{-4mm}
\end{table}%

\noindent\textbf{Comparison with SOTA \short.} In Table~\ref{tab:compare_lvlm_relation}, we compare our \ours~with the state-of-the-art~\short~in the realms of image-based~\cite{zhang2023transfer,gong2023multimodal,ye2023mplugowl,awadalla2023openflamingo,li2023otter}, video-based~\cite{li2023videochat,zhang2023video,Maaz2023VideoChatGPT}, and visually-grounded \short~\cite{chen2023shikra,peng2023kosmos,chen2023minigptv2,wang2023cogvlm,zhao2023svit,zhang2023llavagrounding}, as categorized in Sec.~\ref{sec:related_work}. We selected the MiniGPT4~\cite{zhu2023minigpt} as our baseline due to its straightforward implementation, allowing us to eliminate the confounding effects of more elaborate designs on our experimental results. Notably, the image-based and video-based models lack the capability to output location bounding boxes, precluding the evaluation of their 'BBox Acc' (Bounding Box Accuracy). Moreover, the existing visually-grounded \short~do not support input multiple images simultaneously within one inference process like ours. To enable comparison, we spatially concatenate multiple images into a single one as their input. 

For semantic similarity, our \ours~identifies and localizes the visual commonalities more accurately than other \short. Specifically, \ours~outperforms the second-best method CogVLM~\cite{wang2023cogvlm} by 4.8\% in `BBox Acc' and 0.4 in `RS'.

For semantic contrast, \ours~surpasses the second-best method by a margin of 19.9\% in classification accuracy and 1.4 in `RS', showcasing our model's capability at discerning contrasting relations between images. These results show the superior visual comparison capabilities of \ours.

For temporal associations, \ours~demonstrates best performance on both the SSv2 and ActivityNet test sets, especially excelling at video description tasks on the SSv2 dataset, where it outstrips Video-ChatGPT~\cite{Maaz2023VideoChatGPT}, the second-best method, by 1.3 in `RS'. The SSv2 dataset requires models to concentrate on temporal information, underlining the superior temporal relation comprehension of \ours.

For geometric transformations, \ours~consistently exhibits the best performance, affirming its advanced capabilities in describing the geometric transformations of objects. Additionally, \ours~is capable of tracking object movements with text grounding,  a functionality that the video-based \short~assessed in Table~\ref{tab:compare_lvlm_relation} lack. These multiple capabilities demonstrate the superiority of \ours.

\begin{table}[!htbp]
    \centering
      \caption{Comparison with state-of-the-art (SOTA) person reid methods on describing contrast. `Acc' represents the accuracy of determining if two images are of the same person, 'Prec' is an abbreviation for precision, 'Rec' refers to recall, and 'F1' denotes the F1 score. Meanwhile, 'RS' stands for 'Relation Score,' which quantifies the quality of detailed descriptions.}
        \resizebox{0.5\textwidth}{!}{
        \begin{tabular}{l|cccc|c}
            \toprule
            \multicolumn{1}{c|}{Method} & Acc(\%) & Prec(\%)) & Rec(\%)) & F1(\%)  & RS \\
            \midrule
            ResNet-50 (BOT) & 79.2  & 71.2  & \textbf{97.8}  & 82.4 & - \\
            ResNet-50 (SBS) & 89.8  & 85.4  & 96.0    & 90.4 & - \\
            IBN-50 (MGN) & \textbf{92.2}  & \textbf{91.4}  & 93.2  & \textbf{92.3} & - \\
            \rowcolor[gray]{0.9}\ours~& 83.2  & 77.5  & 93.6  & 84.8 & \textbf{3.3} \\
            \bottomrule
            \end{tabular}%
            }
    \vspace{-4mm}
    \label{tab:compare_expert_contrast}%
\end{table}

Additionally, it is not feasible to align the \short~in Table~\ref{tab:compare_lvlm_relation} for comparison using identical training datasets, as their pre-training and training phases leverage large-scale, general-purpose datasets. These comprehensive datasets already subsume the raw data utilized in the construction of our relation-constrained training data. Furthermore, the data construction critically influences the divergent capabilities of various \short, which also stands as one of our main contributions.

\noindent\textbf{Comparison with task-specific expert models}
In Table~\ref{tab:compare_expert_contrast}, we compare our \ours~with SOTA task-specific expert models on person re-identification, including BoT~\cite{luo2019bag}, SBS~\cite{he2020fastreid}, and MGN~\cite{wang2018learning}. Note that these models cannot provide linguistic answers like \short~but can provide distance measure results, we calculate the distances in the feature space using their released models\footnote{\url{https://github.com/JDAI-CV/fast-reid/blob/master/MODEL_ZOO.md}} for 5,000 positive and negative sample pairs, and use the average distance as the threshold. For the test sample pairs, we classify them as positive samples (\ieno, the same ID) if their feature space distance is less than the threshold, otherwise as negative samples (\ieno, different IDs). Based on the positive and negative sample classification results, we obtain the numerical results for various metrics shown in Table~\ref{tab:compare_expert_contrast}. We observe that our \ours~outperforms the expert model BoT on accuracy and F1 score and achieves comparable results to the other two models, SBS and MGN. Note that these expert models are all trained with task-specific data and loss functions. Moreover, unlike our \ours, they cannot provide natural language descriptions explaining the basis for judgment, \ieno, where the differences lie. These results demonstrate the capability and superiority of our \ours~in visual comparison. We provide more comparison with CLIP-based models on the image retrieval task in supplementary.

\subsection{Ablation Study}
\label{sec:exp_ablation}

We conduct an in-depth ablation study to investigate the impact of different training strategies of \ours. We follow the same evaluation setting proposed in Sec.~\ref{sec:quantitative}, we report the ablation studies in Table~\ref{tab:ablation}.

\noindent\textbf{Ablation of relation data construction.}
The data construction process introduced in Sec.~\ref{sec:data_construction} is pivotal to our approach. It involves extracting predefined visual relations from a series of public datasets to form a unified visual relation dataset. To evaluate our data construction method's influence on performance, an experiment (`w/o data construction' in Table~\ref{tab:ablation}) is conducted where the model is trained on raw data and original labels from public datasets, bypassing the data construction process. This leads to a notable performance drop, as shown in Table~\ref{tab:ablation}, suggesting that our model can handle various visual relation tasks due to the data construction method.

\noindent\textbf{Ablation of LLM finetuning.}
Table~\ref{tab:ablation} indicates that without LLM fine-tuning, performance significantly deteriorates, particularly in bounding box accuracy. Because the special tokens for grounding are not encountered during LLM pretraining, which makes it challenging for the model to learn grounding without fine-tuning the LLM using LoRA. Furthermore, freezing the LLM reduces the number of trainable parameters, resulting in underfitting in complex visual relation tasks.

\noindent\textbf{Vicuna vs. LLaMA-2}
We switch the LLM we used from Vicuna to LLaMA-2~\cite{touvron2023llama2} and observe a slight performance degradation, particularly in traditional metrics. This decrease might stem from the absence of instruction tuning in LLaMA-2, while the Vicuna-based model, having undergone such tuning, follows the instruction better.


\begin{table}[htbp]
  \centering
\caption{The ablation study results, `w.' stands for with, and `w/o' denotes without.}
  \resizebox{\linewidth}{!}{
    \begin{tabular}{l|cc|cc|cc|cc}
    \toprule
    \multicolumn{1}{c|}{\multirow{2}[0]{*}{Method}} & \multicolumn{2}{c|}{Similarity} & \multicolumn{2}{c|}{Contrast} & \multicolumn{2}{c|}{Temporal} & \multicolumn{2}{c}{Geometric} \\
      & \multicolumn{1}{c}{BBox Acc(\%)} & \multicolumn{1}{c|}{RS} & \multicolumn{1}{c}{Acc (\%)} & \multicolumn{1}{c|}{RS} & \multicolumn{1}{c}{RS@SSv2} & \multicolumn{1}{c|}{RS@AN} & \multicolumn{1}{c}{BBox Acc(\%)} & \multicolumn{1}{c}{RS} \\
    \midrule
    w/o data construction & 39.0  & 1.5   & 51.3  & 1.7   & 2.2   & 1.7   & 45.7  & 1.2 \\
    w/o LLM fine-tuning & 30.2  & 2.1   & 67.9  & 1.8   & 2.9   & 1.2   & 32.3  & 1.2 \\
    w. LLaMA-2 & 49.2  & 2.4   & 78.3  & 3.2   & 3.1   & 2.1   & 61.9  & 2.7 \\
    
    \rowcolor[gray]{0.9}\ours  & \textbf{49.3}  & \textbf{2.5}   & \textbf{83.2}  & \textbf{3.3}   & \textbf{3.2}   & \textbf{2.3}   & \textbf{69.7}  & \textbf{3.2} \\
    \bottomrule
    \end{tabular}%
    }
  \label{tab:ablation}%
\end{table}%

\subsection{In-context Learning}
Notably, \ours~exhibits remarkable in-context capability of reasoning from few-shot examples by visually comparison. This demonstrates the generalizability of \ours~on unseen tasks and in real-world scenarios. Comprehensive qualitative and quantitative results are provided in the supplementary materials. Additionally, we synthesize an overview of existing visual in-context learning methods and analyze the underlying reasons for the emergence of in-context learning capabilities in \ours.

\section{Conclusion}

In this paper, we present \ours, a large vision-language model (\shortno) that addresses the limitations of current \short~in their inability of accurately understanding various visual relations, including semantic relations, temporal associations, and geometric transforms. \ours~expands the application scope of \short, especially for tasks that require visual comparisons, taking an important step towards the realization of general-purpose visual understanding system. In more details, we introduce an efficient way to build \ours, including a LLM-powered data construction scheme and a multi-stage model training strategy. The former extracts annotations related to diverse visual relations from existing public datasets and uses the LLM to automatically convert them into a linguistic form for generative training. The latter aims to make full use of the knowledge already acquired by the pre-trained models and further learn how to understand various visual relations based on them, facilitating the learning. Furthermore, we comprehensively evaluate the enabled capabilities of \ours. The qualitative case study results intuitively demonstrate that \ours~can accurately understand diverse visual relations and provide appealing natural language answers. Quantitative comparisons with expert models and existing advanced \short~further prove the superiority of our model in terms of visual comparison capabilities. Besides, we also evaluate the visual in-context learning performance of \ours~and observe favorable generalizing results on unseen tasks, \egno, anomaly detection and medical image recognition. We therefore believe that \ours~has great potential in advancing practical application of \ours~in the near future.

%
%
\bibliographystyle{splncs04}
\bibliography{egbib}

\clearpage
\noindent\textbf{[Supplementary Material]}


In this supplementary material, we demonstrate the in-context learning capabilities of \ours~on novel tasks and in real-world scenarios in Sec.~\ref{sec:in_context_learning}, including analysis of the emergence in Sec.~\ref{sec:why_emergence}, and comprehensive qualitative and quantitative results in Sec.~\ref{sec:in_context_results}. Moreover, we detail the implementation in Sec.~\ref{sec:implementation_details}, including the introduction of datasets and configurations of our multi-stage training strategy in Sec.~\ref{sec:multi_training}, more training prompts details in Sec.~\ref{sec:dialog_prompt}, and details about the evaluation setting in Sec.~\ref{sec:evaluation_details}. We present more experiment results in Sec.~\ref{sec:more_experiments_supp}, including more quantitative results in Sec.~\ref{sec:more_quant_experiments_supp} and more qualitative results in Sec.~\ref{sec:more_quality_experiments_supp}. Additionally, we present more discussion about the potential negative impact, limitations, and code release in Sec.~\ref{sec:more_discussion}.

\section{In-context Learning}
\label{sec:in_context_learning}

In the main paper, we claim that our \ours~exhibits emergent in-context learning capabilities on unseen tasks and real-world scenarios after training on our constructed visual relation data. In Sec.~\ref{sec:why_emergence}, we discuss how visual relations enable the emergence of visual-language in-context learning. And we present both quantitative and qualitative results in Sec~\ref{sec:in_context_results}.

\subsection{Why Emergence?}
\label{sec:why_emergence}
Vision-language in-context learning aims to perform training-free few-shot learning, with a few samples provided as examples. 
Based on different methods of providing examples, we categorize them into three types: 1) \textbf{Language Prompts.} The examples in this kind are provided as purely natural language, augmenting the knowledge of LLMs with new textual contents for unleashing the in-context learning capabilities of LLMs.
In terms of this kind, previous works~\cite{huang2023language,peng2023kosmos} have demonstrated favorable performance. However, relying solely on language prompts may limit their flexibility and applicability. 
2) \textbf{IQA Triplets Examples with Within-triplet Association.} In this category, the provided examples are image-question-answer (IQA) triplets, wherein the visual relationships are depicted within each triplet independently. 
Previous works~\cite{tsimpoukelli2021multimodal,alayrac2022flamingo,li2023otter} enhance the model's capabilities by incorporating this type of examples during training. This method enables the model to learn visual relationships within IQA triplets, generalize to new images and questions, and generate answers. It does not necessitate the recognition of relationships across different images.
3) \textbf{IQA Triplets Examples with Cross-triplet Association.} This category requires the recognition of visual relations across different input example triplets,~\egno~prototype-based classification. The models for this category need to compare the input image with the example images across different triplets finely for getting the final answer. 
This necessitates a robust capability of conducting visual comparison which our~\ours~targets.


Visual relations play a pivotal role in the advancement and emergence of the third category of visual-language in-context learning (\ieno, IQA Triplets Examples with Cross-triplet Association) due to {richer example utilization and contextual relevance. \textbf{Richer Example Utilization:} In-context learning often relies on examples to guide the model. With visual relations, these examples become more informative, providing a scaffold for the model to build upon by comparing the relation between input image and example images. This enriched guidance helps the model to better understand the task at hand and generate more accurate responses. \textbf{Richer Contextual Relevance:} Incorporating visual relations into training enables models to prioritize relevant features and contextual cues. This focus on contextually significant elements allows for more accurate and coherent in-context learning, as the model can draw parallels between related visual and language elements.

\begin{figure*}[!t]
	\centering
	\includegraphics[width=\textwidth]{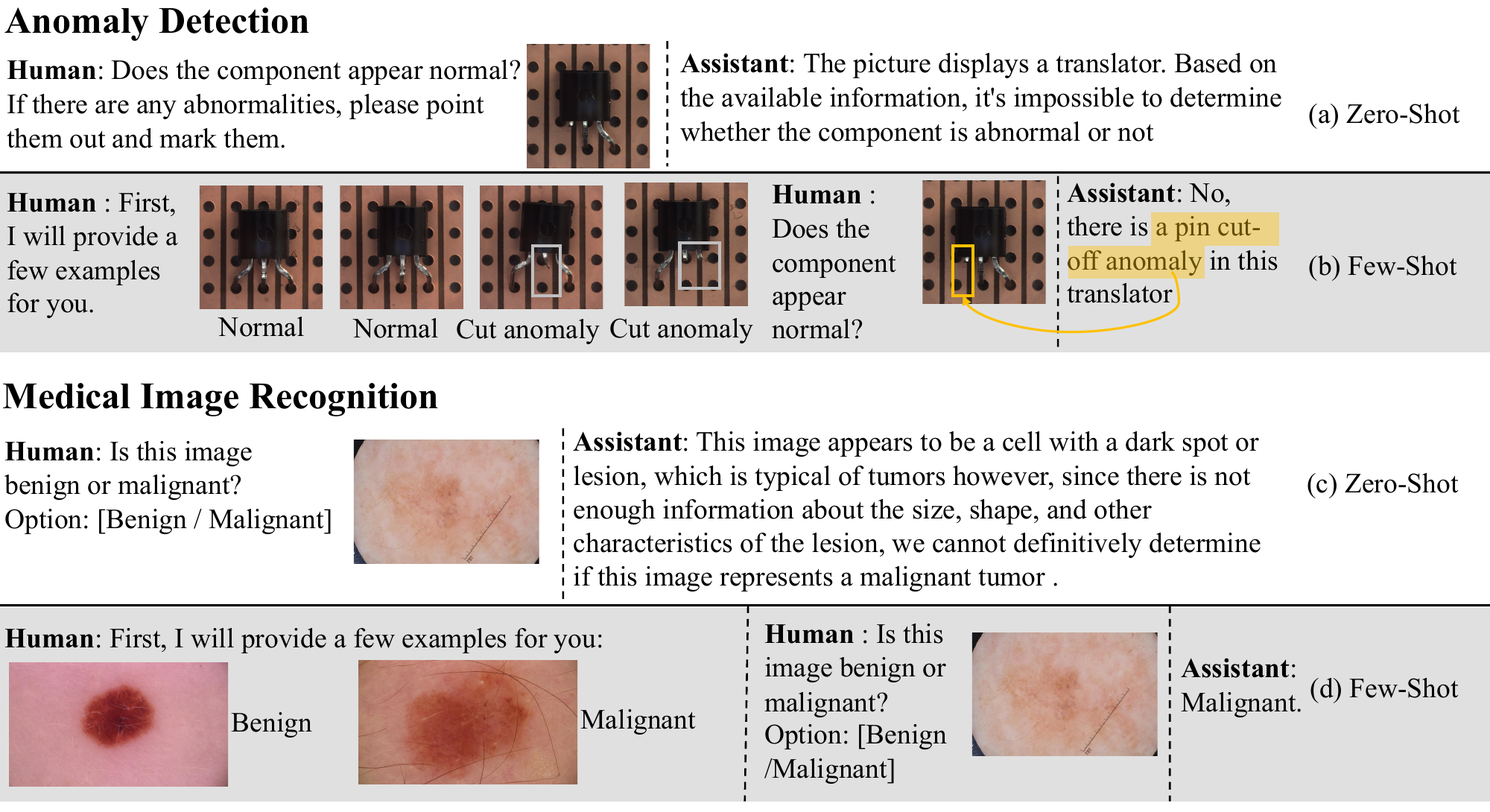}
	\caption{Illustration of RelationVLM's in-context learning performance. (a) The zero-shot (no reference) result on anomaly detection. (b) The few-shot (with few reference images provided) result on anomaly detection. (c) The zero-shot result medical image recognition. (d) The few-shot result on medical image recognition.}
	\label{fig:anomaly_detection}
\end{figure*}

In the following, we demonstrate that our model achieves notable performance in this type of in-context learning tasks outperforming previous approaches.

\subsection{Visual In-context Learning Performance}
\label{sec:in_context_results}

\begin{table}[htbp]
  \centering
\caption{Quantitative results of comparing RelationVLM with SOTA \short~in their visual in-context learning performance. We report the accuracy (\%) of 2/4/8-shot settings on MVTec AD and ISIC datasets for anomaly detection and medical (image) recognition, respectively.}
  \resizebox{0.9\linewidth}{!}{
    \begin{tabular}{l|ccc|ccc}
    \toprule
    \multicolumn{1}{c|}{\multirow{2}[0]{*}{Method}} & \multicolumn{3}{c|}{Acc(\%)@Anomaly Detection}  & \multicolumn{3}{c}{Acc(\%)@Medical Recognition} \\
          & \multicolumn{1}{c}{2-shot} & \multicolumn{1}{c}{4-shot} & \multicolumn{1}{c|}{8-shot}  & \multicolumn{1}{c}{2-shot} & \multicolumn{1}{c}{4-shot} & \multicolumn{1}{c}{8-shot} \\
    \midrule
    LLaVA~\cite{zhang2023transfer} & 54.3  & 53.1  & 53.2  & 54.2  & 53.8  & 53.5 \\
    MM-GPT~\cite{gong2023multimodal} & 53.2  & 54.2  & 54.6  & 52.1  & 52.2  & 52.4 \\
    mPLUG-Owl~\cite{ye2023mplugowl} & 51.7  & 53.7  & 53.9  & 54.5  & 53.9  & 54.1 \\
    Open-Flamingo~\cite{awadalla2023openflamingo} & 60.1  & 62.9  & 64.8  & 55.3  & 57.1  & 57.8 \\
    Otter~\cite{li2023otter} & 62.9  & 65.6  & 67.1  & 54.1  & 60.3  & 61.0 \\
    \hline
    Video-Chat~\cite{li2023videochat} & 52.6  & 52.3  & 50.3  & 53.4  & 52.7  & 51.1 \\
    Video-LLaMA~\cite{zhang2023video} & 50.8  & 52.1  & 52.3  & 50.3  & 51.2  & 51.7 \\
    Video-ChatGPT~\cite{Maaz2023VideoChatGPT} & 51.1  & 51.8  & 52.5  & 51.2  & 52.1  & 52.4 \\
    \hline
    Shikra~\cite{chen2023shikra} & 53.4  & 55.9  & 57.3  & 50.7  & 53.1  & 54.6 \\
    Kosmos-2~\cite{peng2023kosmos} & 54.3  & 54.7  & 59.7  & 54.2  & 55.1  & 55.8 \\
    MiniGPT4-v2~\cite{chen2023minigptv2} & 52.3  & 53.4  & 56.4  & 50.6  & 53.8  & 55.3 \\
    CogVLM~\cite{wang2023cogvlm} & 58.8  & 61.8  & 62.3  & 57.2  & 59.0  & 59.5 \\
    SVIT~\cite{zhao2023svit} & 54.1  & 56.9  & 57.8  & 51.3  & 53.7  & 54.3 \\
    LLaVA-Grounding~\cite{zhang2023llavagrounding} & 54.3  & 58.6  & 59.4  & 55.1  & 57.2  & 57.8 \\
    \hline
    Baseline~\cite{zhu2023minigpt} & 53.9  & 56.2  & 57.6  & 50.9  & 51.1  & 51.2 \\
    \rowcolor[gray]{0.9}\ours (Ours) & \textbf{66.8}  & \textbf{71.1}  & \textbf{72.0}  & \textbf{57.5}  & \textbf{63.6}  & \textbf{65.3} \\
    \bottomrule
    \end{tabular}%
    }
  \label{tab:result_incontext}%
\end{table}%

\noindent\textbf{Settings.} We evaluate the visual in-context learning performance of \ours~on two unseen tasks: anomaly detection and medical image recognition. We use the test data sampled from MVTec AD~\cite{bergmann2019mvtec} and ISIC~\cite{codella2018skin} for them respectively. We present the quantitative comparison results on these two tasks with the advanced \short~in Table~\ref{tab:result_incontext}. Besides, we explicitly illustrate the representative case study results in Figure~\ref{fig:anomaly_detection}.

\noindent\textbf{Quality results.} As shown in Figure~\ref{fig:anomaly_detection} (a), for anomaly detection, \ours~successfully recognizes the object within the given image. When no reference image is provided, our \ours~cannot determine whether the given electronic component in the displayed image is normal or anomalous, but can give a reasonable response to inform the user of what it knows. Once given a small number of reference images, as shown in Figure~\ref{fig:anomaly_detection} (b), our model can precisely identify the specific type of anomaly, which fully demonstrates its excellent visual in-context learning performance on visual comparison. Impressive performance can also be observed on the task of medical image recognition in Figure~\ref{fig:anomaly_detection} (c) (d).

\noindent\textbf{Quantitative results.} Furthermore, we compare our \ours~with advanced \short~in their visual in-context learning performance. As shown in Table~\ref{tab:result_incontext}, when provided with different numbers of reference images, our \ours~consistently exhibits superior in-context learning performance compared to the other two \short. Specifically, \ours~outperforms Otter, the second-best method, by 4.9\% in accuracy for 8-shot settings on the MVTec AD dataset, and by 4.3\% in the 8-shot setting on the ISIC dataset. These experimental results indicate that our \ours~has a stronger ability to understand diverse visual relations and is more adept at performing visual comparison for visual in-context learning.

\section{Implementation Details}
\label{sec:implementation_details}
In this section, we provide implementation details about the multi-stage training (Sec.~\ref{sec:multi_training}), prompts for relation data construction (Sec.~\ref{sec:dialog_prompt}) and the evaluation settings (Sec.~\ref{sec:evaluation_details}).

\begin{table*}[t]
  \centering
    \caption{Details of the datasets used for different training stages.}
  \resizebox{\linewidth}{!}{
    \begin{tabular}{l|l|ccc}
    \toprule
    \multicolumn{1}{c|}{Purposes} & \multicolumn{1}{c|}{Datasets} & Stage-1 & Stage-2 & Stage-3 \\
    \midrule
    Basic Visual Descrip. & LAION~\cite{schuhmann2022laion}, CC3M~\cite{sharma2018conceptual}, SBU~\cite{SBU}. & \checkmark & \checkmark & \checkmark \\
    \hline
    Relation Descrip. & \makecell{GRIT ~\cite{peng2023kosmos}), refCOCO~\cite{yu2016modeling}, CUB-200-2011~\cite{wah2011caltech}, \\ person dataset~\cite{zheng2015scalable,li2014deepreid,xiao2017joint,wei2018person,li2015deepmar,8510891}, MIMIC~\cite{li2023mimicit},\\ SSv2~\cite{goyal2017something}, WebVid~\cite{bain2021frozen}, geometric transforms~\cite{lin2014microsoft}} &       & \checkmark & \checkmark \\
    \hline
    Instruction Tuning & LLaVA-Instruction \cite{liu2023visual}, CCSBU-Aligned \cite{zhu2023minigpt} &       &       & \checkmark \\
    \bottomrule
    \end{tabular}%
    }
  \label{tab:datasets_brief}%
\end{table*}%

\subsection{Multi-Stage Training Strategy} 
\label{sec:multi_training}

We perform three-stage training for facilitating the learning wherein we 
make full use of the pre-trained weights for training \ours~as possible. In all stages, the model is trained using a $224\times224$ image resolution. We employ the AdamW~\cite{loshchilov2017decoupled} optimizer paired with a cosine learning rate scheduler~\cite{loshchilov2016sgdr} for model training. We provide details about training datasets of each stage in Table~\ref{tab:datasets_brief}, and training configurations of each stage in Table~\ref{tab:3stage_training_config}.

\noindent\textbf{Stage-1: basic visual description.} As shown in Table~\ref{tab:datasets_brief}, in stage-1, we warm up the adapter with LAION-5B~\cite{schuhmann2022laion}, CC3M~\cite{sharma2018conceptual}, and SBU~\cite{SBU} for 50k iterations to enable the basic visual description ability of \ours. As shown in Table~\ref{tab:3stage_training_config}, we load the weights for the vision encoder from BLIP-2~\cite{li2023blip} and load the weights of released Vicuna-7B~\cite{vicuna2023} for the LLM-based decoder in \ours. In this stage, we freeze both the vision encoder and the LLM-based decoder. We conduct training over 100,000 steps using 4$\times$A100 GPUs, with a global batch size of 96 and a base learning rate of 2e-3. This stage is completed in approximately 30 hours. 

\noindent\textbf{Stage-2: relation description.} In stage-2, the key stage for training \ours, we continuously train the adapter and fine-tune the LLM-based decoder using 
LoRA~\cite{hu2022lora,peft} on the relation-contained data introduced in the main paper,
with the vision encoder frozen. Specifically, as shown in Table~\ref{tab:datasets_brief}, for semantic relations, we utilize the datasets containing reference expressions and spatial localization information for objects and persons, including GRIT~\cite{peng2023kosmos}, refCOCO~\cite{yu2016modeling}, person reid datasets~\cite{zheng2015scalable,li2014deepreid,xiao2017joint,wei2018person,li2015deepmar,8510891}, CUB-200-2011~\cite{wah2011caltech} and MIMIC~\cite{li2023mimicit}. These datasets inherently carry explicit semantic annotations for entities, enabling the effortless acquisition of labels for semantic relationships between two entities based on their labels. For temporal associations, we use SSv2~\cite{goyal2017something} and WebVid~\cite{bain2021frozen}. The geometric transforms dataset is converted from COCO ~\cite{lin2014microsoft}. We segment natural images and perform geometric transformations (including Horizontal flip, vertical flip, brightness adjustment, rotation, scaling, and moving) on the segmented objects for synthesizing the needed dataset, the corresponding label is the specific transformation applied. Although the dataset is synthesized, we strive to maintain a broad diversity of the synthesized data to enhance the generalization capability of the model as much as possible. Datasets for learning different types of visual relations are jointly used. As shown in Table~\ref{tab:3stage_training_config}. in the second stage, the model is trained for 250,000 steps on 4$\times$A100 GPUs, maintaining a global batch size of 96 and a base learning rate of 2e-4, taking around 80 hours. 

\begin{table}[t]
  \centering
    \caption{Training configuration details on each stage. The snowflake icon represents the freezing of parameters in a particular stage, while the flame icon indicates fine-tuning of parameters during that stage.}
    \resizebox{0.75\textwidth}{!}{
        \begin{tabular}{l|ccc}
        \toprule
        \multicolumn{1}{c|}{Configuration} & Stage1 & Stage2 & Stage3 \\
        \midrule
        \multirow{2}[0]{*}{Vision Backbone} & \snowflake & \snowflake & \snowflake \\
         & BLIP-2 pretrain & BLIP-2 pretrain & BLIP-2 pretrain \\
        \midrule
        \multirow{2}[0]{*}{Adapter} & \flame & \flame & \flame \\
         & From Scratch & From stage1 & From stage2 \\
        \midrule
        \multirow{2}[0]{*}{LLM}   & \snowflake & \flame(LoRA) & \flame(LoRA) \\
         & Vicuna 7b & Vicuna 7b & From stage2 \\
    \midrule
    Base LR & 0.002 & 0.0002 & 0.00002 \\
    Min. LR & 0.00008 & 0.00004 & 0.000008 \\
    Warmup LR & 0.000001 & 0.000001 & 0.000001 \\
    Warm-up scheduler & Linear & Linear & Linear \\
    Scheduler & Cosine & Cosine & Cosine \\
    Weight\_decay & 0.01  & 0.05  & 0.05 \\
    Training iterations & 100000 & 250000 & 10000 \\
    Warmup iterations & 5000  & 10000 & 1000 \\
    Lora\_r & -       & 8     & 16 \\
    Lora\_alpha & -       & 32    & 32 \\
    Lora\_dropout & -       & 0.1   & 0.1 \\
        \bottomrule
        \end{tabular}%
    }
  \label{tab:3stage_training_config}%
\end{table}

\noindent\textbf{Stage-3: instruction tuning.} In stage-3, we perform instruction tuning for the adapter and the LLM-based decoder using LLaVA-instruct-150K dataset~\cite{liu2023visual}, and MiniGPT4 IFT dataset~\cite{zhu2023minigpt}, with the vision encoder still frozen, we manually select some high-quality samples from the training data of the first two stages and combine them with general purpose dialogue data to perform the instruction tuning for VisualCritic, so as to improving the quality and robustness of its responses to user instructions. We select the high-quality subset based on CLIP score, bbox confidence, and answer length. Specifically, we choose data where the CLIP score $>$ 0.34 (if the CLIP score is available), bbox confidence $>$ 0.88 (if bbox confidence is available), and text length $>$ 40 words. As shown in Table~\ref{tab:3stage_training_config}, the model undergoes an additional 10,000 training steps on 4$\times$A100 GPUs, with a global batch size of 64, completed in about 10 hours. The maximum learning rate remains constant at 2e-5 during this final stage.

\subsection{Training Prompts Details} 
\label{sec:dialog_prompt}

\begin{figure*}[!t]
	\centering
	\includegraphics[width=0.75\textwidth]{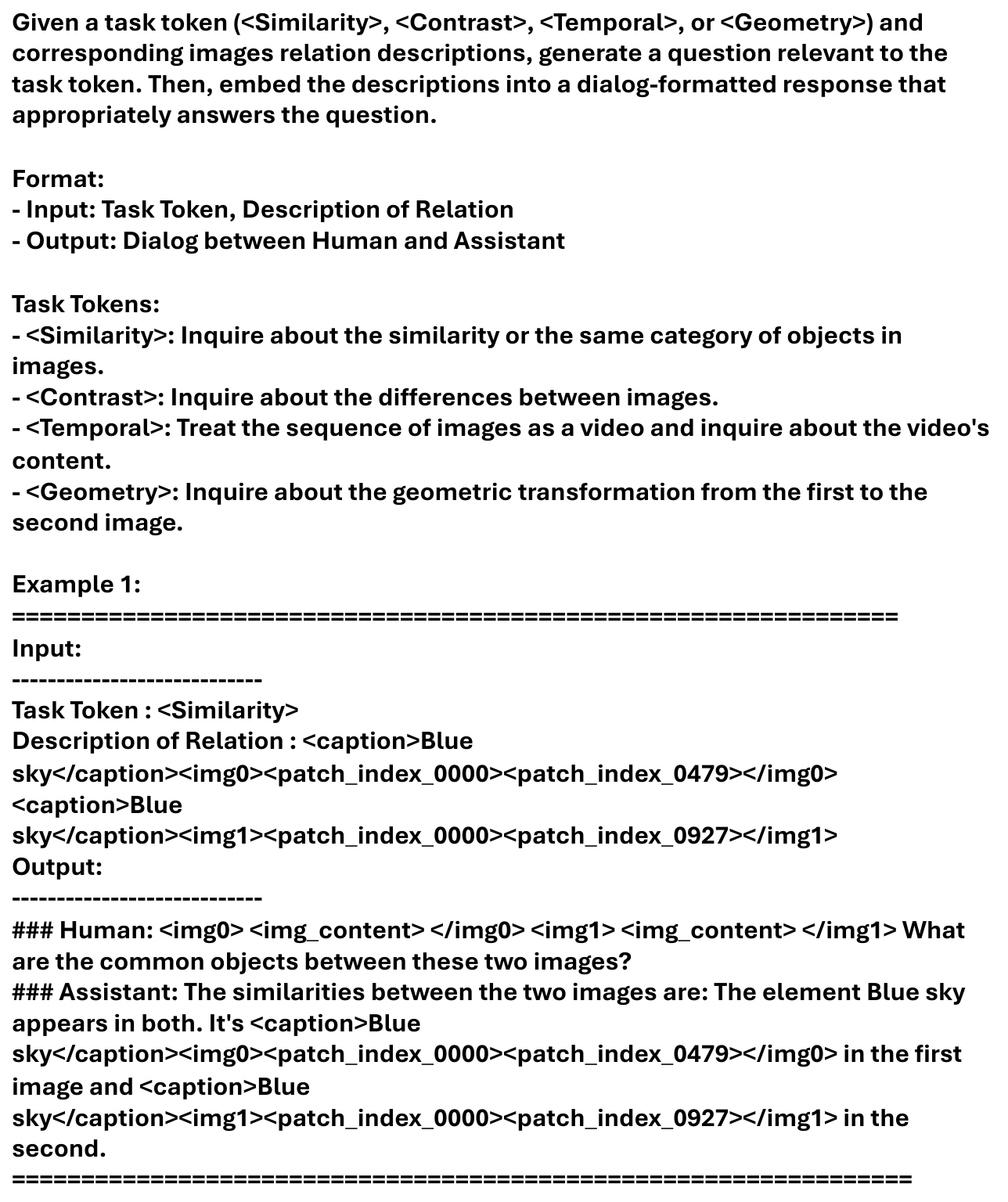}
	\caption{Illustrative dialog prompts for transforming generated relation descriptions into a structured dialogue format, specifically for question-answering, to facilitate generative model training., corresponding to  $p^{dialot}$ in Eq.1 of the main paper.}
	\label{fig:dialog_prompt}
\end{figure*}

In Sec. 3.2 of the main paper, we introduce the process of the relation data construction in detail. As shown in Eq.1 in the main paper, we first collect the image pairs or groups with one or more defined visual relations from the aforementioned datasets. We linguistically encode based on the original annotations present in the source datasets, and then utilize a mature LLM (GPT-4) to automatically generate natural language descriptions for visual relations with description prompts (\ieno, $p^{desc}$ in Eq.1 of the main paper). The description prompts $p^{desc}$ are shown in Fig. 4 of the main paper. Subsequently, we further adopt GPT-4 to further convert linguistic relation descriptions into a dialogue (\ieno, question-answering) form for generative training with dialog prompts (\ieno, $p^{dialog}$ in Eq.1 of the main paper). The dialog prompts are shown in Fig.\ref{fig:dialog_prompt}.

\subsection{Evaluation Details}
\label{sec:evaluation_details}

We show details about the evaluation datasets on Table~\ref{tab:test_datasets}, and the relation score prompt in Fig.~\ref{fig:evaluation_prompt}.

\begin{figure*}[!t]
	\centering
	\includegraphics[width=0.75\textwidth]{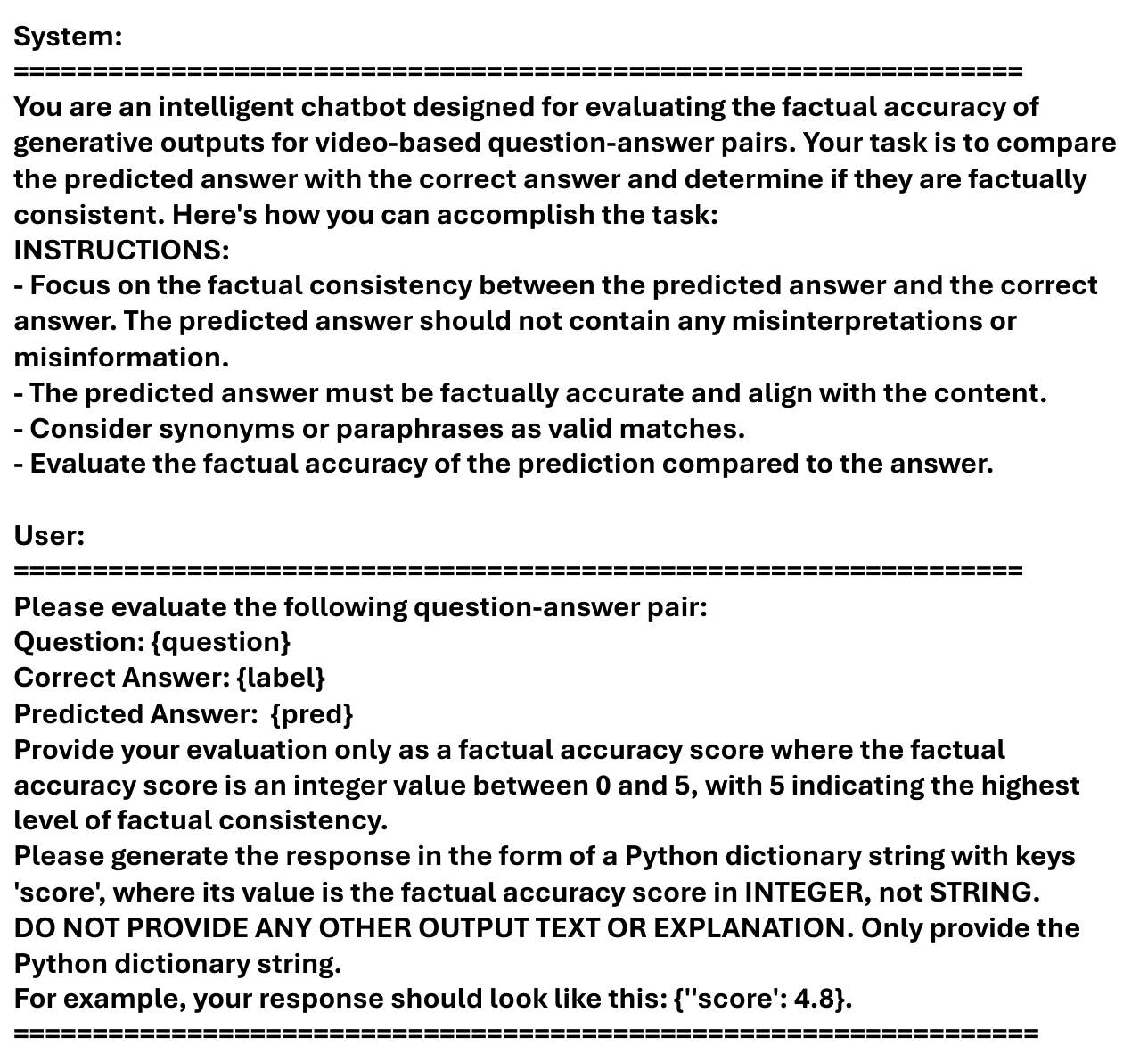}
	\caption{Illustrative evaluation prompts for the Relation Score (\ieno, `RS').}
	\label{fig:evaluation_prompt}
\end{figure*}

\begin{table}[htbp]
    \centering
    \caption{Details of evaluation datasets and question. Using fixed questions stables the test results for better comparison~\cite{li2023evaluating}.}
    \resizebox{.99\linewidth}{!}{
    \begin{tabular}{c|c|c|l}
    \toprule
    Benchmark & Datasets & Number & \multicolumn{1}{|c}{Evaluation Question} \\
    \midrule
    \makecell{Similarity\\ Relation} & COCO  & 5000  & \makecell[l]{What are the common elements or objects found \\in both of these pictures?} \\
    \hline
    \makecell{Contrast\\ Relation} & \makecell{Market1501\\ CUHK03\\ cuhkSYSU\\ MSMT17} & 5000  & Is the same person in these two images? And why? \\
    \hline
    \makecell{Temporal\\ Association} & \makecell{SSv2\\ActivityNet} & 5000  & What is the video about? \\
    \hline
    \makecell{Geometry\\ Transform} & COCO  & 5000  & \makecell[l]{How has the object transformed from the first image \\to the second image?} \\
    \hline
    \multirow{2}[0]{*}{\makecell{In-Context
    \\Learning}} & MVTec AD  & 1000   & Does the component appear normal? \\
     & ISIC  & 1000   & Is this image benign or malignant? option: [benign / malignant] \\
    \bottomrule
    \end{tabular}}
    \vspace{10pt}
    \label{tab:test_datasets}
\end{table}

\begin{table}[!t]
  \centering
  \vspace{-1.5mm}
  \caption{More contrast results on CUB200-2011 dataset. ‘Acc’ represents the accuracy of determining if two images are of the same kind of birds. 'Prec' is an abbreviation for precision, 'Rec' refers to recall, and 'F1' denotes the F1 score. Meanwhile, ’RS’ stands for ’Relation Score,’ which quantifies the quality of detailed descriptions.}
  \resizebox{0.55\textwidth}{!}{
    \begin{tabular}{l|cccc|c}
        \toprule
        \multicolumn{1}{c|}{Method} & Acc(\%) & Prec(\%)) & Rec(\%)) & F1(\%)  & RS \\
    \midrule
    Part RCNN~\cite{zhang2014part} & 74.6 & 69.3 & 88.4 & 77.7 & -  \\
    PS-CNN~\cite{huang2016part} & 76.7 & 71.2 & 90.6 & 79.7 & -  \\
    Bilinear-CNN~\cite{lin2015bilinear} & 85.2 & 80.5 & 93.0 & 86.3  & -  \\
    \hline
    CLIP~\cite{radford2021learning} & 57.8 & 55.4 & 80.0 & 65.5 & - \\
    GIST~\cite{lewis2023gist} & 85.6 & 79.8 & 95.2 & 86.8 &- \\
    \hline
    \rowcolor[gray]{0.9}\ours~ & \textbf{88.8} & \textbf{84.0} & \textbf{95.8} & \textbf{89.5}  & \textbf{2.6} \\
    \bottomrule
    \end{tabular}%
    }
  \label{tab:result_cub}%
\end{table}

\section{More Experiments Results}
\label{sec:more_experiments_supp}

\subsection{More quantitative results}
\label{sec:more_quant_experiments_supp}

\noindent\textbf{Comparing with experts and CLIP Models on fine-grained Images.} In Table~\ref{tab:result_cub}, we present more quantitative results on CUB200-2011~\cite{wah2011caltech} dataset to further evaluate the ability of our RelationVLM in understanding contrast relations. We compare our RelationVLM with some traditional expert models of fine-grained classification task, including Part RCNN~\cite{zhang2014part}, PS-CNN~\cite{huang2016part} and Bilinear-CNN~\cite{lin2015bilinear}, as well as the CLIP-based methods. As these models cannot provide linguistic answers like LVLMs, we employ the same evaluation methods for these models as depicted as ReID models outlined in Table 2 of the main paper. The results, as presented in Table~\ref{tab:result_cub}, substantiate that our RelationVLM not only surpasses in accuracy but also provides linguistic explanations, an attribute absent in the compared models.

\begin{figure*}[!htbp]
	\centering
    \includegraphics[width=0.75\textwidth]{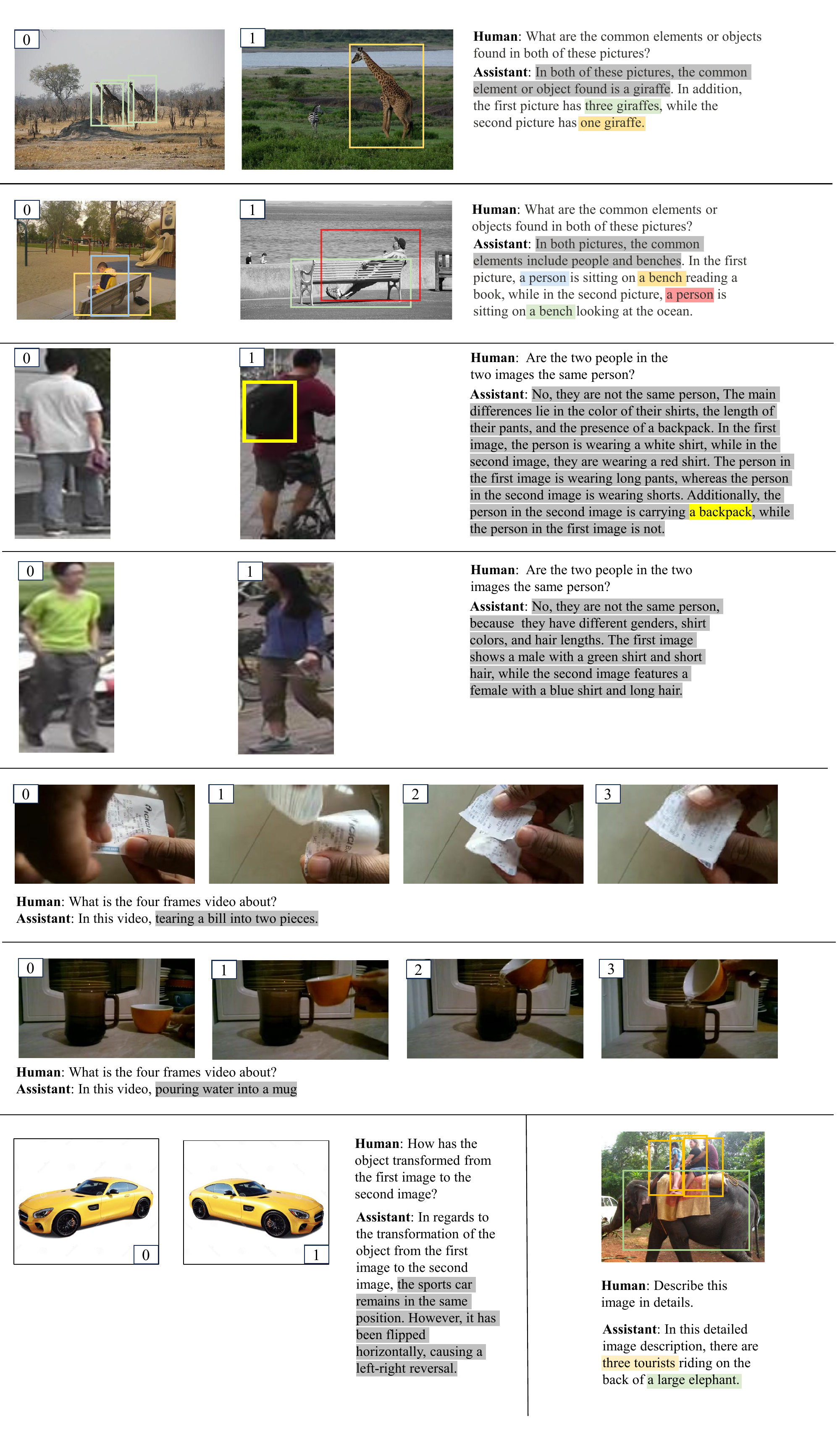}
	\caption{Illustration of more case studies. Color-highlighted text are output along with referring the same color bounding box. And the grey-highlighted text shows details relation description provided by~\ours. And we also showcase a single image QA case here.}
	\label{fig:main_result2}
\end{figure*}

\noindent\textbf{Human side-by-side evaluation.} Quantitative assessment of linguistic outputs from Language-Vision Models (LVMs) remains challenging. Commonly, a more sophisticated LVM, such as GPT-4, is utilized to critique and evaluate the performance of other models. We report the `Relation Score', \ieno, `RS' following this method. Additionally, we implemented a human-centric comparative analysis involving 10 participants who assessed our model against four other \short~on 100 random relation description samples. The G+S/S+B scores (where \textbf{G}ood: our VisualCritic preferred, \textbf{S}ame: no preference, \textbf{B}ad: other model preferred) are as shown in Table~\ref{tab:side-by-side}.

\vspace{-0.9em}
\begin{table}[!h]
\caption{Human side-by-side comparison of ours with other \short~in qualitative Results.}
\centering
\resizebox{0.9\linewidth}{!}{
\begin{tabular}{l|cccc}
\toprule
    \multicolumn{1}{c|}{G+S/S+B} & GPT-4V~\cite{gpt4vision} & LLaVA-v1.5~\cite{liu2023visual} & MiniGPT-4~\cite{zhu2023minigpt} & InstructBLIP~\cite{instructblip}  \\ 
    \midrule
    \rowcolor[gray]{0.9}VisualCritic (Ours) v.s. & 1.14 & 1.93 & 1.83 & 2.67 \\
    \bottomrule
\hline
\end{tabular}
}
\vspace{-1.2em}
\label{tab:side-by-side}
\end{table}

\subsection{More quality results}
\label{sec:more_quality_experiments_supp}
More cases in which~\ours~describes the relation are in Figure~\ref{fig:main_result2}. And we also show a single image QA case, showing our model's ability to answer the question about a single image.



\section{More Discussion}
\label{sec:more_discussion}

\noindent \textbf{Potential negative impact.} For resource intensity, the training and inference of \ours~often require substantial computational resources, which could have environmental impacts due to energy consumption and contribute to the carbon footprint. For data bias, the model may inadvertently perpetuate or amplify biases present in the training data constructed from original biased public datasets. This could result in discriminatory outcomes if the model is used in applications.

\noindent \textbf{Limitation.} One of our current main limitations lies in the data used for handling geometric transform is synthesized by us, rather than being collected from the real world. We have not yet found any publicly annotated data that can be utilized, so we synthesize them. We will further improve our model in this aspect in the future study. Additionally, the diversity and bias of our data depend to some extent on GPT-4, which is recognized as the most advanced large-scale model. In fact, evaluating the performance of large-scale models, including the diversity and bias of their data, remains a highly challenging issue. 

\noindent \textbf{Code release.} Our data, code, and models will be released upon paper acceptance.

\end{document}